\documentclass[10pt,letterpaper]{article}

\usepackage[left=2.40cm, right=2.40cm, top=2.40cm, bottom=2.40cm]{geometry}
\usepackage[numbers]{natbib}

\usepackage{amsmath}
\usepackage{amsfonts}
\usepackage{amsthm}
\usepackage{graphicx}
\usepackage{caption}
\usepackage{subfig}
\usepackage{framed}
\usepackage{multicol}

\usepackage[titletoc,title]{appendix}
\usepackage{algorithm}
\usepackage{algpseudocode}

\DeclareMathOperator*{\argmin}{arg\,min}
\newcommand{\bydef}{\stackrel{\mathrm{def}}{=}}

\usepackage{mathtools}

\title{Local Information with Feedback Perturbation Suffices for Dictionary Learning in Neural Circuits}
\author{
	Tsung-Han Lin\\ \\
	Intel Labs \\
	Santa Clara, CA, USA\\
	tsung-han.lin@intel.com     
}

\date{}

\numberwithin{equation}{section}

\begin{document}

\maketitle
\begin{abstract}
	
While the sparse coding principle can successfully model information processing in sensory neural systems, it remains unclear how learning can be accomplished under neural architectural constraints.
Feasible learning rules must rely solely on synaptically local information in order to be implemented on spatially distributed neurons.
We describe a neural network with spiking neurons that can address the aforementioned fundamental challenge and solve the $\ell_1$-minimizing dictionary learning problem, representing the first model able to do so.
Our major innovation is to introduce feedback synapses to create a pathway to turn the seemingly non-local information into local ones.
The resulting network encodes the error signal needed for learning as the change of network steady states caused by feedback, and operates akin to the classical stochastic gradient descent method.


\end{abstract}

\section{Introduction}
A spiking neural network (SNN) is a computational model with simple neurons as the basic processing units.
Different from artificial neural networks, SNNs incorporate the time dimension into computations.
The network of neurons operates according to a global reference clock; at a time instance, one or more neurons may send out a 1-bit impulse, the \textit{spike}, to neighbors through directed connectivities, known as \textit{synapses}. 
Neurons form a dynamical system with local state variables and rules that determine when a neuron transmits a spike.
The spike rate of a neuron can encode its activation value, borrowing the terminology from artificial neural networks.

\begin{figure}[t]
	\center
	\begin{tabular}{cc}
		\begin{tabular}{c}
			\includegraphics[scale=0.33]{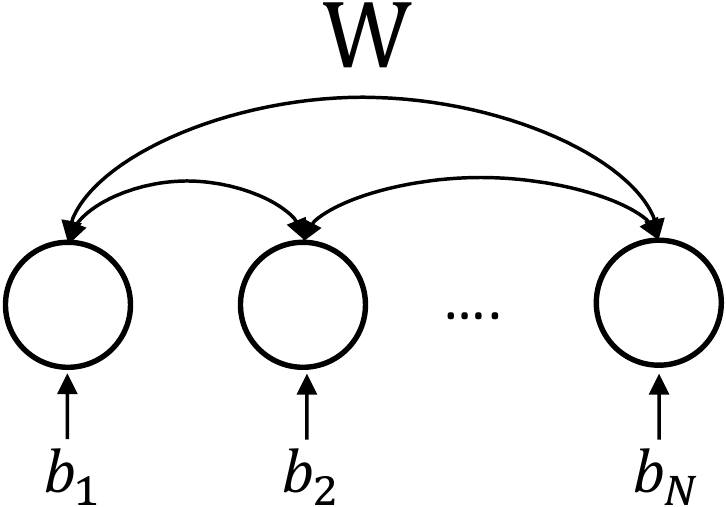} \\ (a) Laterally connected network
		\end{tabular}
		&
		\begin{tabular}{c}
			\includegraphics[scale=0.33]{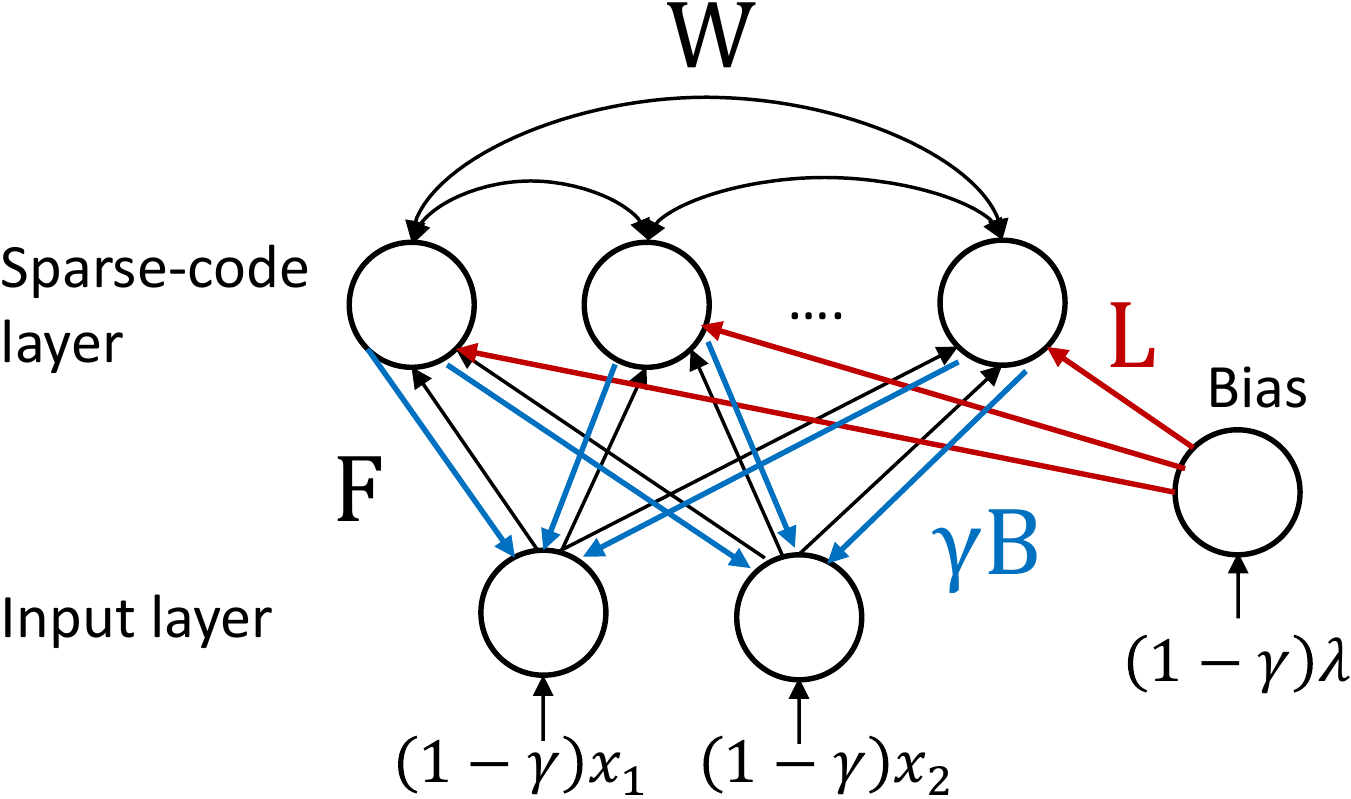} \\ (b) Dictionary learning network
		\end{tabular}
	\end{tabular}
	\caption{The two network topology for sparse coding and dictionary learning.
		Our main focus is the network in (b).
		The firing thresholds of the input layer and bias neurons are set to 1.	
	}
	\label{fig:networks}
\end{figure}

SNNs can exploit the temporal ordering of spikes to obtain high computational efficiency, despite that encoding real values as spike rates may appear quite inefficient comparing to the compact binary representations.
Consider, for example, a set of competing neurons recurrently connected with inhibitory synapses in Figure \ref{fig:networks}(a).
The winner neuron that has the largest external input will fire at the earliest time, and immediately inhibit the activities of other neurons.
This inhibition happens with only a single one-to-many spike communication, in contrast to the all-to-all state exchange and comparison required when neurons only maintain graded activation values.  
Using the above principle, one can show that a SNN can be configured to efficiently solve the well-known $\ell_1$-minimizing sparse approximation problem \cite{shapero2014optimal, tang2016sparse}, which is to determine a sparse subset of features from a feature dictionary to represent a given input, and the features can be viewed as competing neurons that seek to form the best fit of the input data \cite{rozell2008sparse}. 

In this work, we further show that the related dictionary learning problem can be solved in a SNN as well. 
Dictionary learning was first proposed to model mammalian visual cortex \cite{olshausen1996emergence}, and later found numerous applications in image processing and machine learning \cite{mairal2014sparse}. 
Despite its popularity, it remains unclear how the problem can be solved in a neural architecture.
None of the existing learning algorithms are synaptically local: the adaptation of synaptic weights relies on the receptive field information of other neurons, making it impossible to be implemented in a spatially distributed network. 
As a result, many researchers turn to other less straightforward objective function formulations (e.g., minimizing over long-term average neuron activities \cite{zylberberg2011sparse}, or maximizing input-output similarity \cite{hu2014hebbian}), or are forced to take approximate gradient directions at the cost of suboptimal results (e.g., simplifying the learning rules to be only Hebbian \cite{brito2016nonlinear,vertechi2014unsupervised}).

We solve the dictionary learning problem by introducing \textit{feedback synapses}.
We show that the feedback connections can cause the network steady states to change by an amount identical to the error signal needed for learning, provided that the network synaptic weights satisfy a weight consistency condition.
Built on this observation, we develop learning mechanisms that closely resemble the classical stochastic gradient descent, and can perform dictionary learning from a spiking network with randomly initialized synaptic weights. 

\section{Integrate-and-Fire Neuron Model}
\label{sec:network_dynamics}

We first consider a network of $N$ simple integrate-and-fire neurons.
Each neuron-$i$, for $i=1,2,\hdots,N$, has two internal state variables, the \textit{soma current} $\mu_i(t)$ and the \textit{membrane potential} $v_i(t)$, that govern its dynamics. 
The soma current is determined by two inputs: The first is a constant current $b_i$; the second are the
spike trains of all the neighbors neuron-$j$s to which neuron-$i$ is connected. Each spike train is of the form
 $\sigma_j(t) = \sum_k \delta(t-t_{j,k})$ where $t_{j,k}$ is the time of the $k$-th spike of neuron-$j$ and $\delta(t)$ is the Dirac delta
function. The soma current $\mu_i(t)$ is the sum of $b_i$ and the filtered spike trains from its neighbors thus:
\begin{equation}
\mu_i(t) = b_i + \sum_{j\neq i} w_{ij}(\alpha \ast \sigma_j)(t)
\end{equation}
where $w_{ij}$ is the synaptic weight from neuron-$j$ to $i$, and $\alpha(t) = \tau_s^{-1} H(t)e^{-t/\tau_s}$ is the filter kernel parameterized by the synaptic decay time constant $\tau_s$; $H(t)$ is the Heaviside function that is 1 when $t \geq 0$ and 0 elsewhere.
Note that in a neural architecture, synaptic weights are stored, and hence only available, at the destination neuron. This property is referred to as synaptically local, and constitutes the major challenge for dictionary learning in SNNs.

The soma current is converted to the output spiking activities through the dynamics of membrane potential. This membrane potential is a simple linear integration of the soma current before it reaches the firing threshold.
\begin{equation}
v_i(t) = \int_{t_{i,k}}^{t} \mu_i(s) ds 
\end{equation}
A spike is generated when the membrane potential exceeds its firing threshold $\theta_i$; at this time, the neuron also immediately resets $v_i$ to 0. 

The network of neurons forms a dynamical system where the neurons interact through spikes.
An important quantity $\Delta_i$, called the \textit{imbalance function}, is useful in characterizing the steady states of the system, defined for $t>0$ as follows,
\begin{equation}
\Delta_i(t) \bydef u_i(t) - \theta_i a_i(t)
\end{equation}
where $u(t)$ and $a(t)$ are the average soma current and average spike rate, respectively,
\begin{equation}
u_i(t) \bydef \frac{1}{t} \int_{0}^{t} \mu_i(s) ds 
,\;\;\;\;\;\;\;\;
a_i(t) \bydef \frac{1}{t} \int_{0}^{t} \sigma_i(s) ds 
\end{equation}
The imbalance function measures the difference between the average amount of charges accumulated to the membrane potential, equal to $u_i(t)$, and the average amount of charges released through spiking, equal to $\theta_i a_i(t)$.

If the average current converges to a fixed point, one can show that as $t \rightarrow \infty$, the imbalance converges towards satisfying the following \textit{equilibrium condition} \cite{tang2016sparse},
\begin{equation}
\begin{aligned}
\left\{\begin{matrix}
\lim_{t \rightarrow \infty} \Delta_i(t) = 0, &\;\;\textrm{ if } \lim_{t \rightarrow \infty} a_i(t) > 0 &\\
\lim_{t \rightarrow \infty} \Delta_i(t) \leq 0, &\;\;\textrm{ if } \lim_{t \rightarrow \infty} a_i(t) = 0 & \\
\end{matrix}\right.
\end{aligned}
\label{eq:equi_rate}
\end{equation}
This result simply suggests that for neurons that have nonzero spike rates at equilibrium, their average outgoing charges must equal the average incoming charges, meaning the imbalance must be 0.
On the contrary, if a neuron stops spiking at equilibrium, then the imbalance may either be zero, if it has zero net incoming charges, or a negative value, if it receives more inhibition than excitation. 
We note that the convergence property of the dynamical system deserves a rigorous treatment (e.g., see \cite{shapero2014optimal,tang2016sparse}), although it is not the main focus of this work.

\section{Nonnegative Sparse Coding}
Solving the sparse coding problem under a given dictionary constitutes an important part of our learning scheme.
In this section, we revisit prior results on solving this problem in a SNN \cite{shapero2014optimal,tang2016sparse}, and we will focus on using nonnegative dictionaries.

Consider the network topology in Figure \ref{fig:networks}(a). Each neuron-$i$ receives an input current $b_i$, and has incoming synapses with weight $w_{ij}$ from the other $(N-1)$ neurons. Suppose that the synapses are all inhibitory, that is, $w_{ij} \leq 0$, and hence none of the neurons can spike arbitrarily fast. 
Using (\ref{eq:equi_rate}), the equilibrium spike rates $a_i^* = \lim_{t \rightarrow \infty}a_i(t)$ must satisfy 
\begin{align}
\left\{\begin{matrix}
&b_i + \sum_{j\neq i} w_{ij} a_j^* - \theta_i a_i^* = 0, &\textrm{ \;\;\; if } a_i^* > 0
\\ 
&b_i + \sum_{j\neq i} w_{ij} a_j^* \leq 0, &\textrm{ \;\;\; if } a_i^* = 0
\end{matrix}\right.
\label{eq:snn_balance}
\end{align}
The above result makes use of the property that the average current will converge to 
$\lim_{t \rightarrow \infty}u_i(t) = b_i + \sum_{j\neq i} w_{ij}a_i^*$.

The steady-state condition, (\ref{eq:snn_balance}), has connections to the nonnegative sparse coding problem,
\begin{equation}
\mathbf{a}^* = \argmin_\mathbf{a \geq 0} \frac{1}{2}\left \| \mathbf{x - Da} \right \|^2_2 + \lambda \sum_j \left(\sum_i d_{ij}^2\right) a_j 
\label{eq:nn_lasso}
\end{equation}
where $\mathbf{x} \in \mathbb{R}^M$ is a data sample, $\mathbf{D} \in \mathbb{R}^{M \times N}$ is a dictionary, and $\lambda \geq 0$ is the sparse regularization parameter. 
To see the connection, let $\mathbf{G} = \mathbf{D}^T \mathbf{D}$ and $g_{ij}$ be its $(i,j)$-th entry. 
The necessary and sufficient optimality condition for (\ref{eq:nn_lasso}) is
\begin{align}
\left\{\begin{matrix}
&\sum_k d_{ki} x_k - \lambda g_{ii} - \sum_{j\neq i} g_{ij} a_j^* - g_{ii} a_i^* = 0,&\textrm{ \;\;\; if } a_i^* > 0
\\ 
&\sum_k d_{ki} x_k - \lambda g_{ii} - \sum_{j\neq i} g_{ij} a_j^* \leq 0, &\textrm{ \;\;\; if } a_i^* = 0
\end{matrix}\right.
\label{eq:lasso_optimality}
\end{align}
Note the similarity between (\ref{eq:snn_balance}) and (\ref{eq:lasso_optimality}).
The correspondence can be established by setting
\begin{equation}
b_i = \sum\nolimits_k d_{ki} x_k  - \lambda g_{ii}, \;\;\;\;
w_{ij} = -g_{ij}, \;\;\;\;
\theta_i = g_{ii}
\label{eq:nnls_snn_config}
\end{equation}
Indeed, previous work has established that a spiking network configured as above will converge to an equilibrium spike rate identical to the solution of (\ref{eq:nn_lasso}). 
Note that the dictionary is not encoded explicitly in the network. The input current is configured according to the dictionary projection of the input data, and the synaptic weights represent the correlations between columns of the dictionary.

\section{Online Dictionary Learning}
We are interested in learning a nonnegative dictionary from $P$ nonnegative training data samples, $\mathbf{x}_1, \mathbf{x}_2, \hdots, \mathbf{x}_P$. 
The dictionary learning problem is commonly formulated as,
\begin{align}
\begin{split}
\min_{\mathbf{D}\geq 0,\left \| \mathbf{d}_j\right \|_2^2 = 1,\mathbf{a}_i \geq 0} \frac{1}{P}\mathbf \sum_{i=1}^{P} L\left(\mathbf{x}_{i}, \mathbf{D},\mathbf{a}_i\right), \;\;\; 
L\left(\mathbf{x}, \mathbf{D},\mathbf{a}\right)  = \frac{1}{2}\left \| \mathbf{x} - \mathbf{Da} \right \|^2_2 + \lambda \left \|\mathbf{a} \right\|_1
\end{split}
\label{eq:dict_learning}
\end{align}
where $\mathbf{a}_i$ is the sparse representation of data $\mathbf{x}_i$.
The number of columns, or \textit{atoms}, in the dictionary, $N$, is a predetermined hyper-parameter.
This optimization problem seeks the best performing dictionary for all data samples, minimizing the sum of all sparse coding losses.


\subsection{A Two-Layer Network for Dictionary Learning}

Consider the network topology in Figure \ref{fig:networks}(b) that consists of two layers of neurons, an input layer of $M$ neurons at the bottom and a sparse-code layer of $N$ neurons on top.
There are four groups of synaptic weights: the excitatory feedforward and feedback synapses, $\mathbf{F} \in \mathbb{R}^{N \times M}$, $\gamma \mathbf{B} \in \mathbb{R}^{M \times N}$ where $\gamma$ is a scalar in [0,1), and $\mathbf{F}, \mathbf{B} \geq 0$; the inhibitory lateral and bias synapses, $\mathbf{W} \in \mathbb{R}^{N \times N}$, $\mathbf{L} \in \mathbb{R}^{N \times 1}$, and $\mathbf{W}, \mathbf{L} \leq 0$. 

In the case of $\gamma=0$, Figure \ref{fig:networks}(b) is an instantiation of Figure \ref{fig:networks}(a), where the constant current inputs are replaced by spike trains of identical averages.
To see this, note that the feedback synapses are removed in this setting, and the input and bias neurons will spike at a constant rate $x_1,x_2,\hdots,x_M$ and $\lambda$, as they are only driven by constant external inputs.
We can similarly establish the correspondence between the equilibrium at the sparse-code layer, as in (\ref{eq:snn_balance}), and the optimality condition for sparse coding in (\ref{eq:lasso_optimality}), by configuring the network as follows,
\begin{equation}
	f_{ij} = d_{ji}, \;\;\;\; 
	w_{ij} = -g_{ij}, \;\;\;\; 
	\theta_i = g_{ii}, \;\;\;\; 
	l_i = -\theta_i
\label{eq:dict_learn_ff}
\end{equation}
$\theta_i$ being the firing threshold of neuron-$i$ in the sparse-code layer.
From (\ref{eq:dict_learn_ff}), we can see that dictionary learning in this network means adapting the feedforward weights towards the optimal dictionary, and the lateral weights towards the correlations between the optimal dictionary atoms.
In addition, learning proceeds in an online manner, where data samples are given sequentially by swapping inputs, and the dictionary is updated as soon as a new data sample is available. 

We derive learning mechanisms that resemble the classical online stochastic gradient descent \cite[Sec 5.5]{mairal2014sparse}, consisting of two iterative steps. 
The first step computes the optimal sparse code with respect to the current dictionary, and the second step updates the dictionary by 
estimating the gradient from a single training sample, giving the following update sequence
\begin{equation}  
\begin{aligned}
\mathbf{D}_{i+1} 
&= \Pi_\mathcal{C} \left[ \mathbf{D}_{i} - \eta \nabla_{\mathbf{D}} L\left(\mathbf{x}_i, \mathbf{D}_{i}, \mathbf{a}^*_i\right) \right]\\
&= \Pi_\mathcal{C} \left[ \mathbf{D}_{i} - \eta \left(\mathbf{D}_{i}\mathbf{a}^*_i - \mathbf{x}_i\right) \mathbf{a}^{*T}_i \right]
\end{aligned}
\label{eq:algo_recon_error}
\end{equation}
where the projection operator $\Pi_\mathcal{C}$ projects to the positive quadrant and renormalizes each atom in the updated dictionary, and $\eta$ is the learning rate.

We operate the spiking network with two stages to mimic the above two iterative steps.
In the first stage, called the \textit{feedforward stage}, we set $\gamma = 0$ and feed the training sample $\mathbf{x}_i$ to the input layer neurons. From the discussions above, the optimal sparse code can be found as the equilibrium spike rates at the sparse-code layer.
The main challenges lie in the second stage where the dictionary needs to be updated using synaptically local mechanisms, whereas the information needed appears to be non-local for the following two reasons:
\textbf{1) Reconstruction error cannot be locally computed at the sparse-code layer.}
The gradient consists of a reconstruction error term, which is crucial to determining the best way to adapt the dictionary. Unfortunately, computing $\mathbf{D}\mathbf{a}^*$ requires the full knowledge of $\mathbf{D}$, but only one column of the dictionary is local to a sparse-code neuron.
\textbf{2) Atom correlations are non-local to compute.}
The lateral synaptic weights should be updated to capture the new correlations between the updated atoms.
Again, computing a correlation requires knowledge of two atoms, while only one of them is accessible by a sparse-code neuron.
In the next section, we show how feedback synapses can be exploited to address these two fundamental challenges.

\subsection{Synaptically Local Learning}
\label{sec:ff_rules}

\paragraph{Reconstruction with Feedback Synapses.}
In the second stage of learning, called \textit{feedback stage}, we set $\gamma$ to a nonzero value to engage the feedback synapses, moving the network towards a new steady state.
Interestingly, there exists a condition that if satisfied, engaging the feedback synapses will only perturb the equilibrium spike rates at the input layer, while leaving the sparse-code layer untouched. 
We call this condition \textit{feedback consistency},
\begin{equation}
\mathbf{FB} = \mathbf{H}, \textrm{\;\; where \;\;} \mathbf{H} =
\begin{bmatrix}
\theta_1 & -w_{12} & \hdots & -w_{1N}\\ 
-w_{21} & \theta_2 &  & -w_{2N} \\ 
\vdots &  & \ddots  & \\ 
-w_{N1} & -w_{N2}  &  & \theta_N
\end{bmatrix} 
\label{eq:fb_consistency}
\end{equation}
Note that $\mathbf{H}$ is composed of the lateral weights $w_{ij}$ and firing thresholds $\theta_i$. 

To see this, let $\mathbf{y} \in \mathbb{R}^{M}$, $\mathbf{z} \in \mathbb{R}^{N}$ be the equilibrium spike rates at the input and sparse-code layer, respectively. 
The equilibrium spike rates at the input layer can be easily derived.
Given that the input neurons do not receive any inhibition, their imbalance functions must be zero at equilibrium, and hence their spike rates are,
\begin{equation}
\begin{aligned}
\mathbf{y}^{(1)} &= \mathbf{x} \\
\mathbf{y}^{(2)} &= (1-\gamma) \mathbf{x} + \gamma\mathbf{B} \mathbf{z}^{(2)} \\
\end{aligned}
\label{eq:input_equilibrium}
\end{equation}
with superscripts denoting the particular learning stage that the equilibrium spike rates belong to.

For the sparse-code layer, note that the equilibrium spike rate must satisfy the equilibrium condition in (\ref{eq:equi_rate}). This allows us to examine the relationships between $\mathbf{z}^{(1)}$ and $\mathbf{z}^{(2)}$. Let $\mathbf{e} \in \mathbb{R}^{N}$ be the imbalance of the sparse-code layer neurons at equilibrium, $e_i = \lim_{t \rightarrow \infty} \Delta_i(t)$, we can write
the imbalance during the feedforward and feedback stage at equilibrium as
\begin{equation}
\mathbf{e}^{(1)} = \mathbf{Fx} + \lambda\mathbf{L} - \mathbf{Hz}^{(1)} 
\label{eq:ff_equilibrium}
\end{equation}
\begin{equation}
\begin{aligned}
\mathbf{e}^{(2)} 
&= \mathbf{Fy}^{(2)} + (1-\gamma)\lambda\mathbf{L} - \mathbf{Hz}^{(2)} \\
\end{aligned}
\label{eq:fb_equilibrium0}
\end{equation}
Substituting (\ref{eq:input_equilibrium}) into (\ref{eq:fb_equilibrium0}),
\begin{equation}
\begin{aligned}
\mathbf{e}^{(2)} = (1-\gamma) \mathbf{F}\mathbf{x} + \gamma\mathbf{F}\mathbf{B}\mathbf{z}^{(2)} + (1-\gamma)\lambda\mathbf{L} - \mathbf{Hz}^{(2)} \\
\end{aligned}
\label{eq:fb_equilibrium1}
\end{equation}
Now, suppose that the feedback weights satisfy feedback consistency, $\mathbf{FB}=\mathbf{H}$, $\mathbf{e}^{(2)}$ can be further reduced,
\begin{equation}
\begin{aligned}
\mathbf{e}^{(2)} &= (1-\gamma) \left(\mathbf{Fx} + \lambda\mathbf{L} - \mathbf{Hz}^{(2)} \right)\\
\end{aligned}
\label{eq:fb_equilibrium}
\end{equation}
Note the similarity between (\ref{eq:ff_equilibrium}) and (\ref{eq:fb_equilibrium}), and $1-\gamma > 0$.
This suggests that a feasible $\mathbf{z}^{(1)}$ that satisfies (\ref{eq:equi_rate}) must also be a feasible $\mathbf{z}^{(2)}$, and vice versa.
In other words, if the feedforward-only network possesses a unique equilibrium spike rate, then the sparse-code layer spike rates must remain unaltered between the two learning stages, $\mathbf{z}^{(2)} = \mathbf{z}^{(1)}$.

With this result, we turn our attention to the amount of spike rate changes at the input layer in a feedback-consistent network.
\begin{equation}
\mathbf{y}^{(2)} - \mathbf{y}^{(1)} = \gamma(\mathbf{B} \mathbf{z}^{(1)} - \mathbf{x}) \\
\end{equation}
As the sparse code computed in the feedforward stage is preserved in the feedback stage, the change amounts to a reconstruction error, in that the reconstruction is formed by the feedback weights $\mathbf{B}$ as the dictionary.
Suppose for now that the feedforward and feedback weights are symmetric (except for a scalar factor $\gamma$), consistent, and equal to an underlying dictionary $\mathbf{D}$, that is, $\mathbf{B}=\mathbf{F}^T=\mathbf{D}$ and $\mathbf{H}=\mathbf{D}^T\mathbf{D}$, which is also the ideal situation that learning should achieve.
The reconstruction errors needed in the gradient calculations become locally available as the change of input layer spike rates.
This leads to the following synaptically local learning rules that update the weights along the desired gradient direction in (\ref{eq:algo_recon_error}),
\begin{equation}
\begin{split}
f_{ij} \leftarrow f_{ij} + \eta_f z_{i}^{(2)} \left(y_j^{(1)} - y_j^{(2)}\right) - \lambda_f f_{ij}  \\
b_{ij} \leftarrow b_{ij} + \eta_b \left(y_i^{(1)} - y_i^{(2)}\right) z_j^{(2)} - \lambda_b b_{ij}
\end{split}
\label{eq:ff_learning_rule}
\end{equation}
$\eta_f$ and $\eta_b$ being the learning rates.
Note that a weight decay term is included at the end to prevent the weights from growing too large, with $\lambda_f$ and $\lambda_b$ being the regularization coefficients.
This is where our algorithm departs from the classical stochastic gradient descent, as renormalizing the atoms in the feedback weights is non-local.
In addition, we truncate the weight values when they go below zero to ensure their nonnegativity.

In the case of asymmetric weights, we can still adopt the learning rules above.
Initially, the weight updates may not be able to improve the dictionary, given that the reconstruction in the feedback stage is formed using a dictionary quite different from the encoding dictionary $\mathbf{F}$.
However, over many updates, the weights will gradually become symmetric, since the learning rules adjust both feedforward and feedback weights in the same direction, and their initial differences will diminish with the decay term.
When the two weights become sufficiently aligned, the learning rules will likely find a descending direction, despite not the steepest, towards the optimal dictionary.
Perfect symmetry is not necessary for learning to work. 

\paragraph{Maintaining Feedback Consistency.}

Feedback consistency is the key property behind the rationale of the learning mechanism above. 
Suppose that a network is initialized to be feedback-consistent, after an update to its feedforward and feedback weights, one must adjust the lateral weights and firing thresholds accordingly to restore the consistency.
Unfortunately, direct computations, $\mathbf{H} \leftarrow \mathbf{F}\mathbf{B}$, are not synaptically local.
The sparse-code layer neurons, who can modify $\mathbf{H}$, do not have access to the feedback weights, which are local to the input layer neurons.

To avoid non-local computations, we instead have the sparse-code layer neurons minimize the following inconsistency loss $L_c$ given $P_s$ training samples, again using stochastic gradient descent
\begin{equation}
\min_{\mathbf{H}}\frac{1}{P_s}\sum_{i=1}^{P_s} L_c(\mathbf{H}, \mathbf{z}_i^{(2)}) \;\;\;\;,\;\; L_c(\mathbf{H}, \mathbf{z}) = \frac{1}{2} \left \| (\mathbf{H-FB}) \mathbf{z} \right \|^2
\label{eq:inconsistency_loss}
\end{equation}
The key observation is that the inconsistency loss can be measured from the difference in equilibrium spike rates of sparse-code neurons between the two learning stages.
Their relationship can be easily shown by reorganizing (\ref{eq:ff_equilibrium}) and (\ref{eq:fb_equilibrium1}), 
\begin{equation}
\begin{aligned}
\gamma(\mathbf{H-FB}) \mathbf{z}^{(2)}
&= -\mathbf{e}^{(2)} + (1-\gamma) \mathbf{e}^{(1)} - (1-\gamma)\mathbf{H} \left( \mathbf{z}^{(2)} - \mathbf{z}^{(1)} \right)
\end{aligned}
\end{equation}
We can then derive the gradient to minimize $L_c$
\begin{equation}
\begin{aligned}
\nabla_{\mathbf{H}} L_c (\mathbf{H}, \mathbf{z}_i^{(2)})
&= (\mathbf{H-FB})\mathbf{z}^{(2)} \left(\mathbf{z}^{(2)}\right)^T \\
&= \frac{1}{\gamma} \left( -\mathbf{e}^{(2)} + (1-\gamma) \mathbf{e}^{(1)} - (1-\gamma)\mathbf{H} \left( \mathbf{z}^{(2)} - \mathbf{z}^{(1)} \right) \right)  \left(\mathbf{z}^{(2)}\right)^T
\end{aligned}
\end{equation}
Note that the gradient above can be computed with synaptically local information.
Suppose for now that $\mathbf{F}$ and $\mathbf{B}$ are fixed during which the sub-problem is being solved, we then can use the rule $\mathbf{H} \leftarrow \mathbf{H} - \eta_h\nabla_{\mathbf{H}} L_c - \lambda_h\mathbf{H}$ to update both lateral weights and firing thresholds. 
With a sufficiently large $P_s$, feedback consistency can be restored.

We can further relax the assumption that $\mathbf{F}$ and $\mathbf{B}$ are fixed during which $\mathbf{H}$ is adjusted, by using a much faster learning rate, $\eta_h \gg \eta_f$ and $\eta_h \gg \eta_b$.
In other words, $\mathbf{F}$ and $\mathbf{B}$ are approximately constant when the network is solving (\ref{eq:inconsistency_loss}).
All learning rules then can be activated and learn simultaneously when a new training sample is presented.
The network eventually will learn an underlying dictionary $\mathbf{D} = \mathbf{F}^T \approx \mathbf{B}$, and the optimal lateral weights $\mathbf{H} \approx \mathbf{D}^T\mathbf{D}$.

\section{Numerical Simulations}

We examined the proposed learning algorithm using three standard datasets in image processing, machine learning, and computational neuroscience.
\textbf{Dataset A.} Randomly sampled $8\times 8$ patches from the grayscale Lena image to learn 256 atoms.  
\textbf{Dataset B.} $28\times 28$ MNIST images \cite{lecun1998gradient} to learn 512 atoms.
\textbf{Dataset C.} Randomly sampled $16\times 16$ patches from whitened natural scenes \cite{olshausen1996emergence} to learn 1024 atoms.
For Dataset A and C, the patches are further subtracted by the means, normalized, and split into positive and negative channels to create nonnegative inputs \cite{hoyer2004non}.
The spiking networks are ran with a time step of $1/32$. 
For each input, the feedforward stage is ran from $t=0$ to $t=20$ and the feedback stage is ran from $t=20$ to $t=40$, and the spike rates are measured simply as the total number of spikes within the time window of 20.
We deliberately chose a short time window (the spike rates only have a precision of 0.05) to demonstrate the fast convergence of spike patterns; a more accurate equilibrium spike rate may be obtained if one is willing to use a larger window starting at some $t>0$.
The synaptic weights are randomly initialized to be asymmetric and inconsistent, with the lateral weights set to be sufficiently strong so that the spike rates will not diverge in the feedback stage. 
For the learning rates, we set $\eta_f = \eta_b$ and $\eta_h = 32\eta_f$.

\paragraph{Learning Dynamics.}  

\begin{figure}[t]
	\centering
	\begin{tabular}{cc}
		\begin{tabular}{c}
			\includegraphics[scale=0.26]{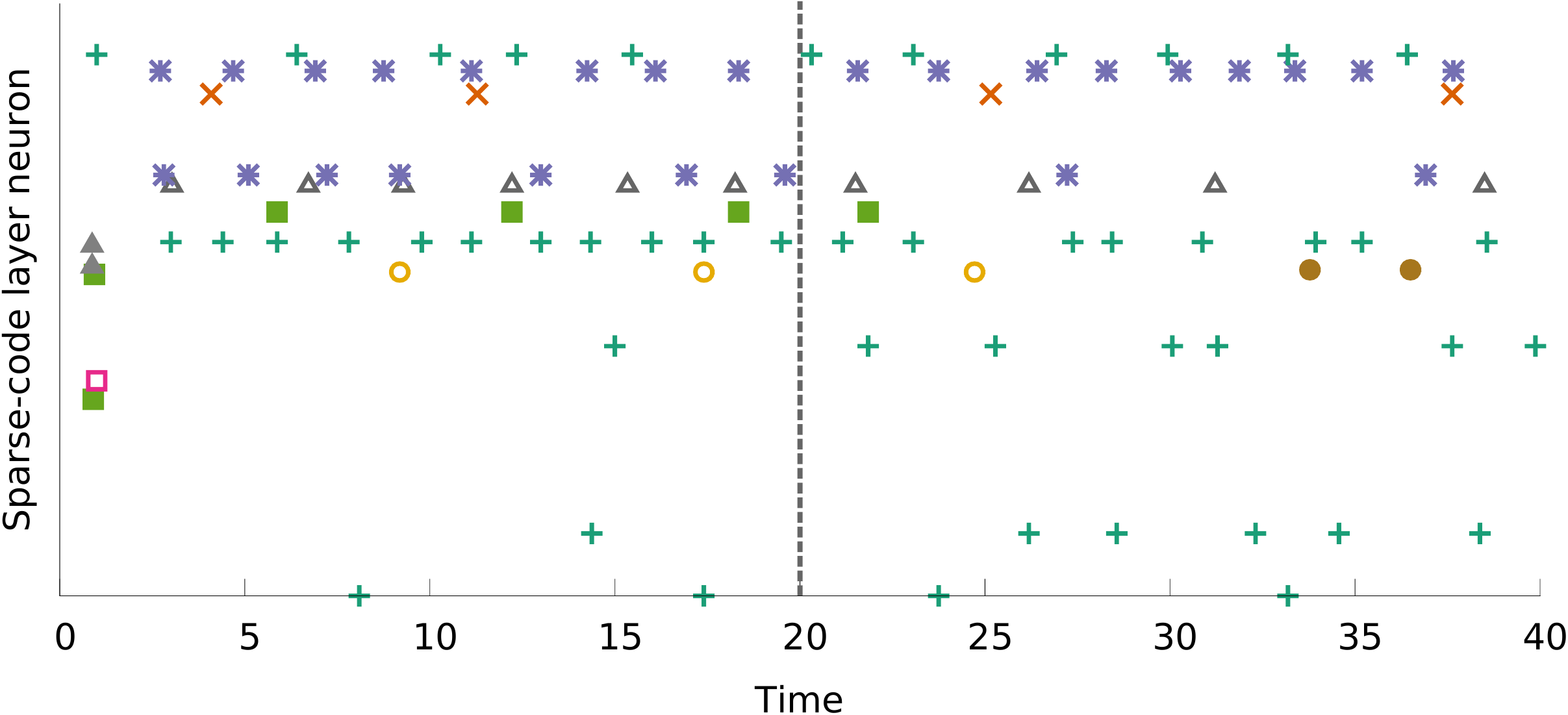} 
		\end{tabular}
		&
		\begin{tabular}{c}
			\includegraphics[scale=0.26]{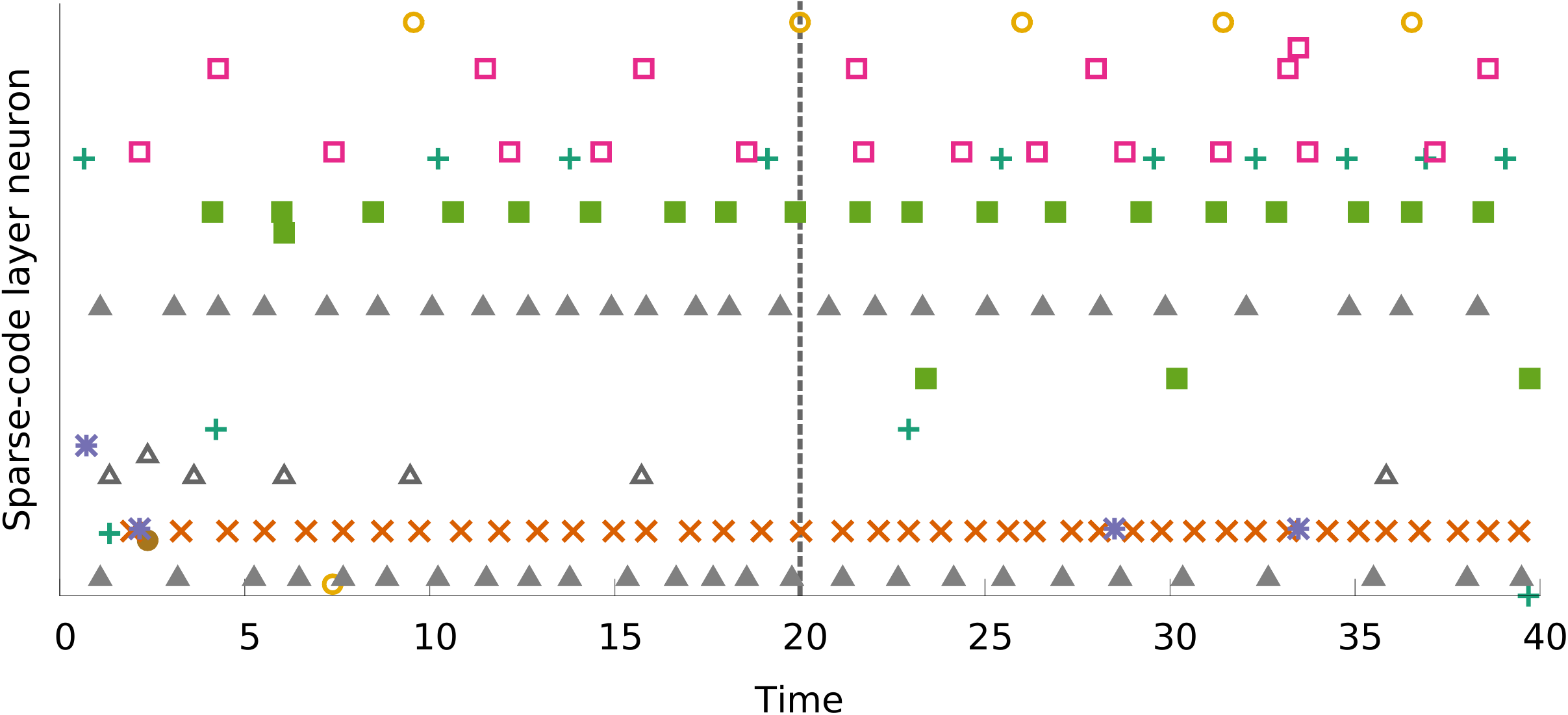} 
		\end{tabular} 
		\\
		\begin{tabular}{c}
			\includegraphics[scale=0.26]{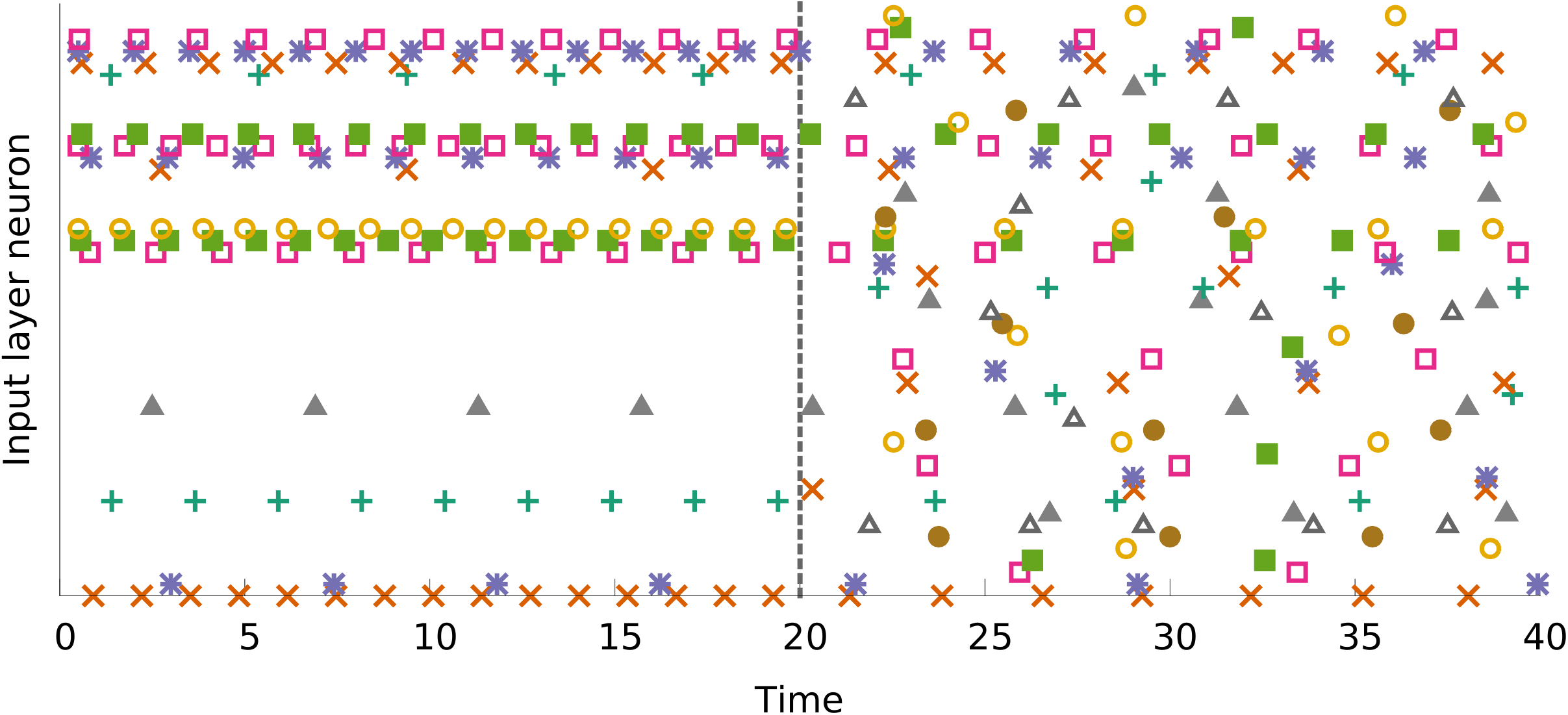} 
		\end{tabular}
		&
		\begin{tabular}{c}
			\includegraphics[scale=0.26]{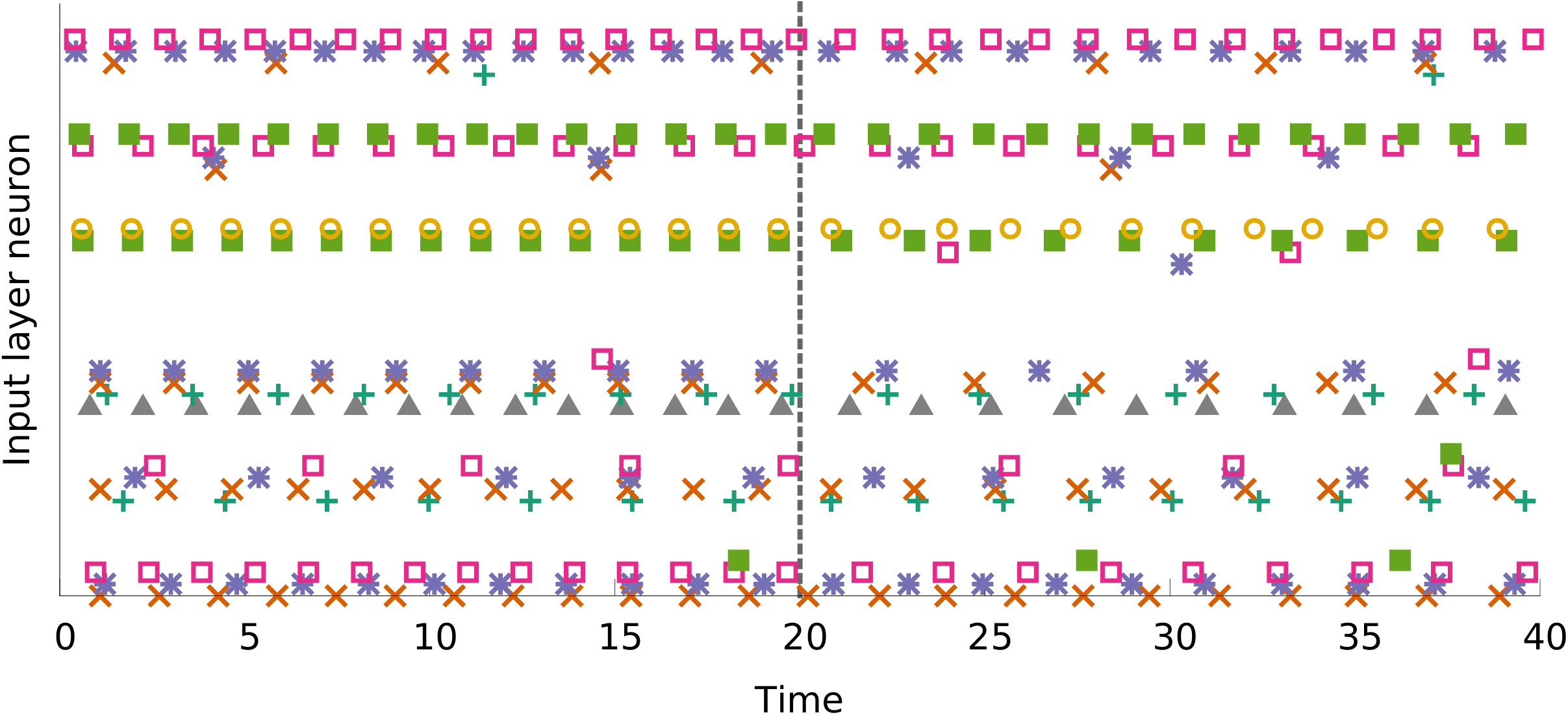} 
		\end{tabular}		
	\end{tabular}
	\caption{Network spike patterns with inputs from Dataset A. \textbf{Left panel:} before learning. \textbf{Right panel:} after learning.  The top panel are the spike rasters of sparse-code layer neurons, and the bottom panel are the rasters of input layer neurons. Only a subset of the input layer is shown. }
	\label{fig:network_dynamics}
\end{figure}

\begin{figure}[t]
	\centering
	\begin{tabular}{cc}
		\begin{tabular}{c}
			\includegraphics[scale=0.13]{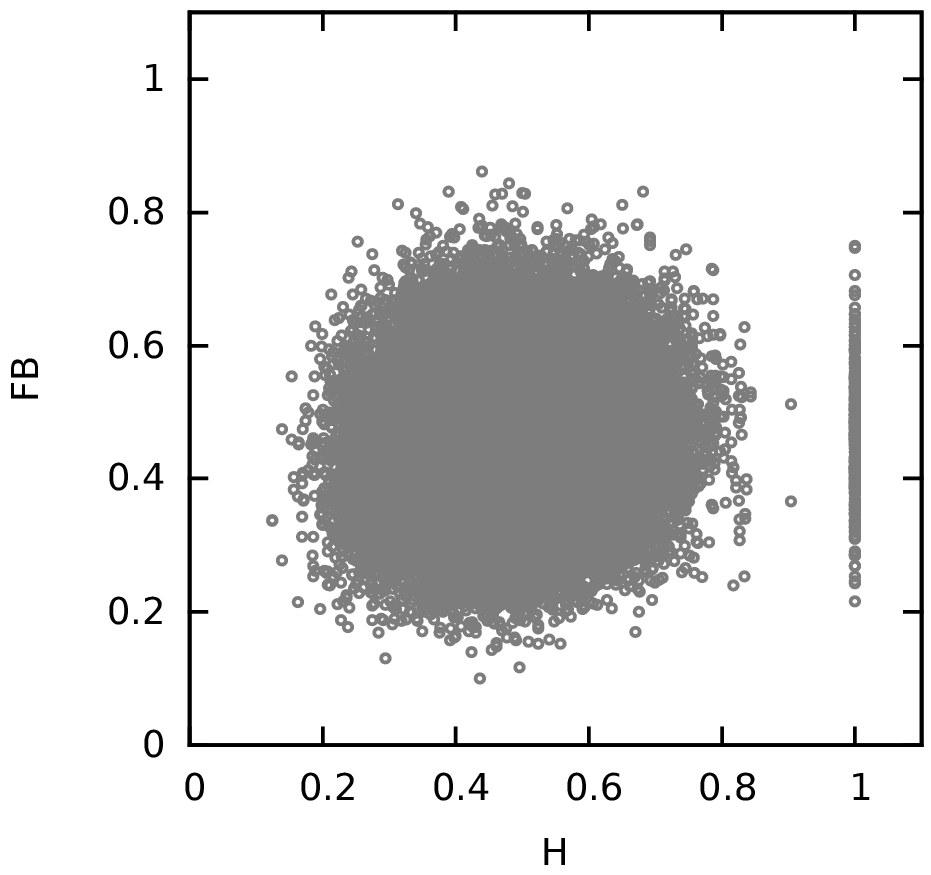} 
		\end{tabular}
		&
		\begin{tabular}{c}
			\includegraphics[scale=0.13]{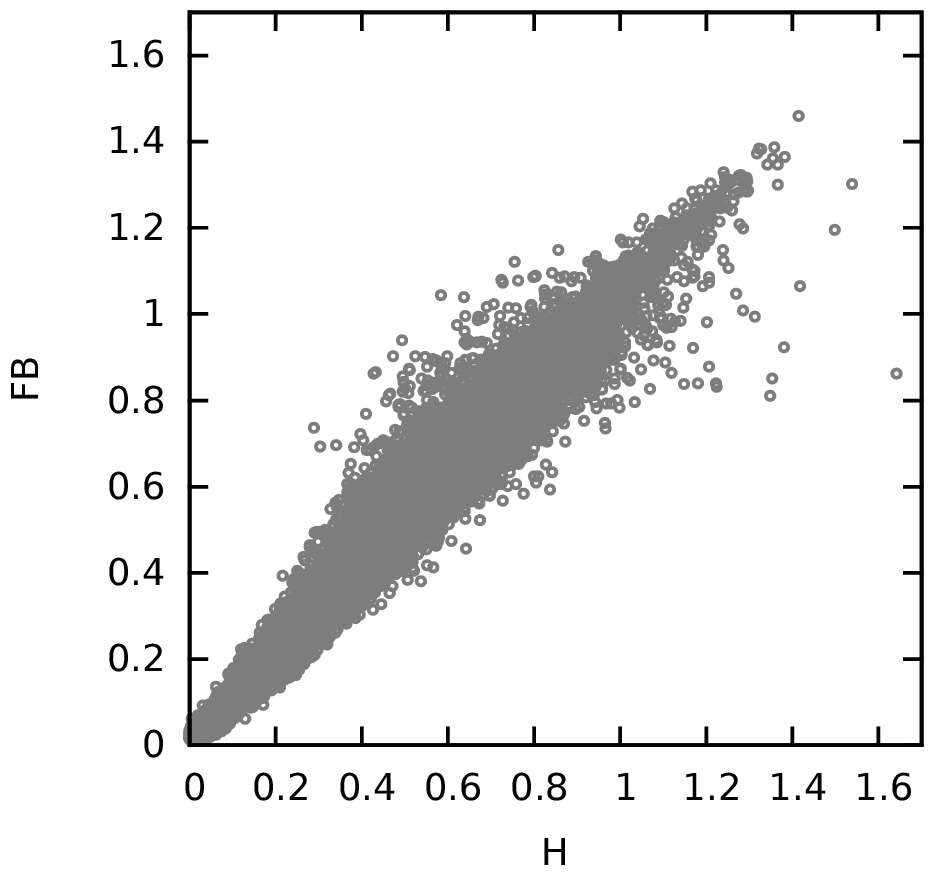} 
		\end{tabular}
	\end{tabular}
	\caption{The network learns to satisfy feedback consistency. Figures show the scatter plot of $\mathbf{H}$ versus $\mathbf{FB}$. \textbf{Left:} before learning; the vertical line corresponds to the initial firing thresholds all set to 1. \textbf{Right}: after learning.}
	\label{fig:lateral_weights}
\end{figure}

Figure \ref{fig:network_dynamics} shows the spike patterns before and after learning in both layers.
Before learning, we see both sparse-code and input layer neurons exhibit perturbed spike rates in the feedback stage, as predicted in the earlier section. 
The perturbation in sparse-code neurons is caused by the inconsistency between randomly initialized synaptic weights, while the perturbation in input neurons is additionally due to the large reconstruction errors.
After learning, the spike patterns become much steadier as the network learns to maintain weight consistency and minimize reconstruction error.
Figure \ref{fig:lateral_weights} shows the scatter plot of the learned lateral weights and firing thresholds, $\mathbf{H}$, versus their desired values, $\mathbf{FB}$. 
It can be seen that after learning, the network is able to maintain feedback consistency.


\paragraph{Comparison with Stochastic Gradient Descent.} 

Dictionary learning is a notorious non-convex optimization problem.
Here we demonstrate the proposed algorithm can indeed find a good local minimum. We compare the convergence behavior with stochastic gradient descent (SGD) with batch size of 1, to which our algorithm closely resembles. 
For SGD, we use the same learning rate as the spiking network, $\eta = \eta_f$, and explore two nearby learning rates $\eta = 2 \eta_f$ and $\eta = 0.5\eta_f$.
Additionally, we experiment initializing the spiking network weights to be symmetric and consistent to understand the impact of random initialization.
The weight decay rates are chosen so that the firing thresholds, which correspond to the squared norms of atoms, converge to a dynamic equilibrium around 1 to ensure a fair comparison. 
For each dataset, a separate test set of 10,000 samples is extracted, whose objective function value is used as the quality measure for the learned dictionaries. 

\begin{figure}[t]
	\centering
	\includegraphics[scale=0.39]{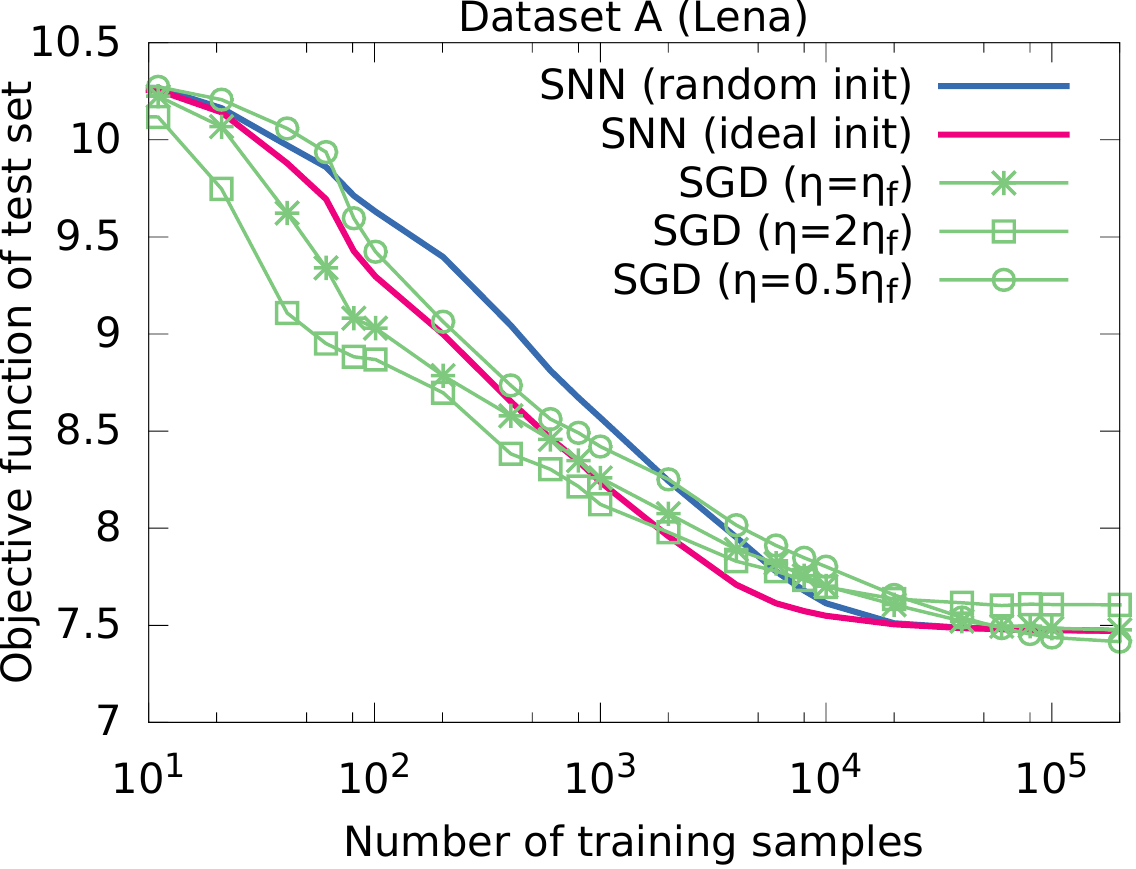} 
	\includegraphics[scale=0.375]{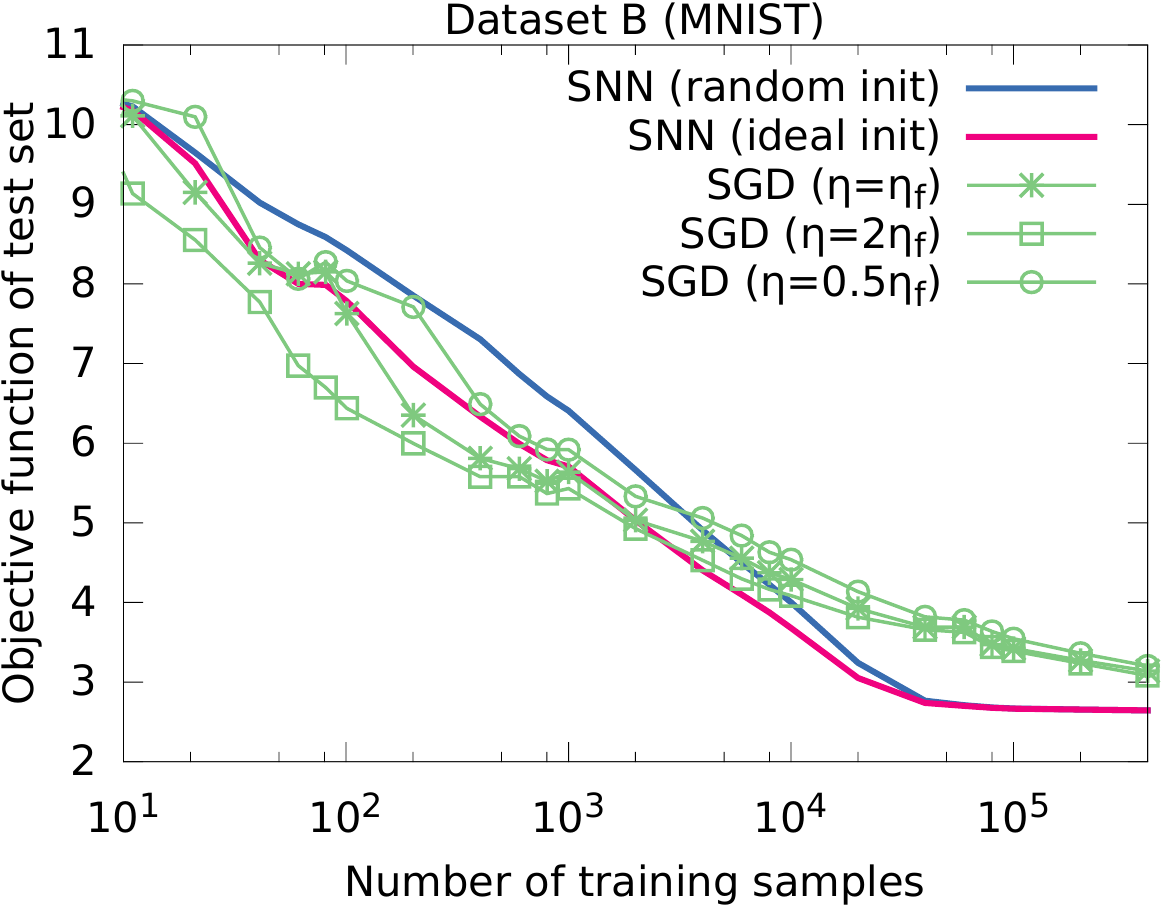} 
	\includegraphics[scale=0.375]{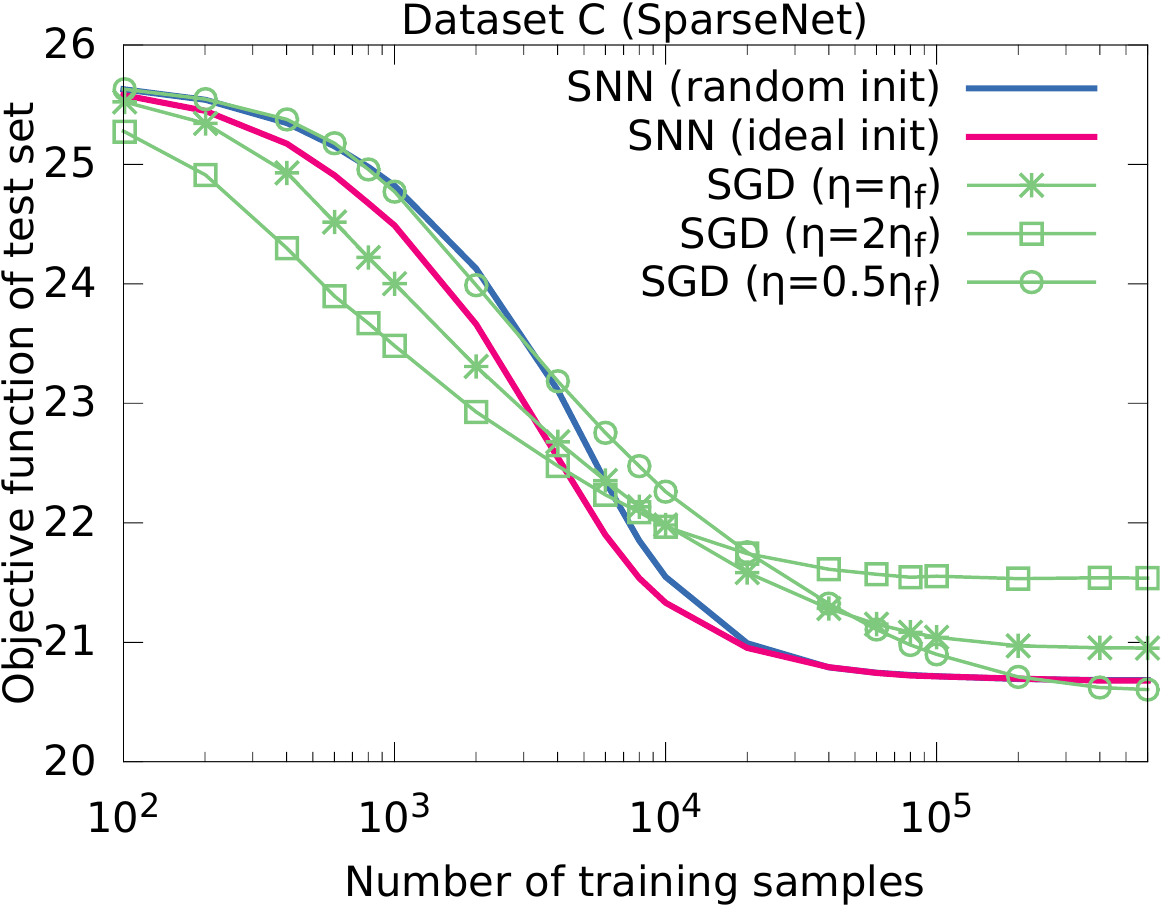} 
	\\
	\includegraphics[scale=0.75]{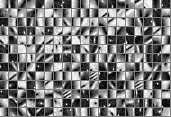} 
	\includegraphics[scale=0.235]{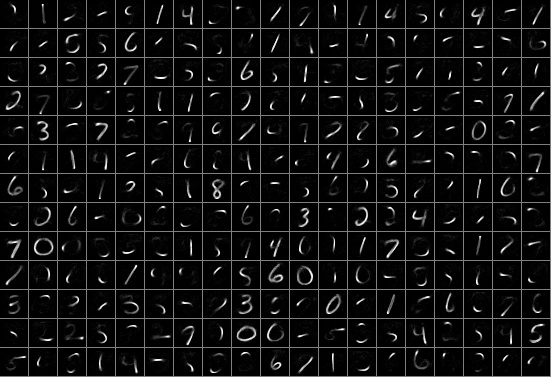} 
	\includegraphics[scale=0.4]{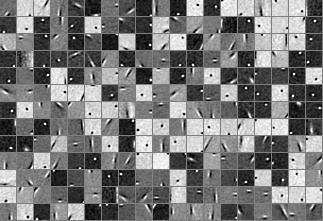} 		
	\caption{Comparison of SNN and stochastic gradient descent. The bottom panel shows a random subset of the dictionaries learned in SNN. They show patterns of edges and textures (Lena), strokes and parts of the digits (MNIST), and Gabor-like oriented filters (natural scenes). }
	\label{fig:convergence}
\end{figure}

Figure \ref{fig:convergence} shows that our SNN algorithm can obtain a solution of similar, if not better, objective function values to SGD consistently across the datasets. 
Surprisingly, the SNN algorithm can even reach better solutions with fewer training samples, while SGD can be stuck at a poor local minimum especially when the dictionary is large.
This can be attributed to the dynamic adaptation of firing thresholds that mitigates the issue in SGD that some atoms can be rarely activated and remain unlearned.
In SNN, if an atom is not activated over many training samples, its firing threshold decays, which makes it more likely to be activated for the next sample.
Further, we observe that random weight initialization in SNN only causes slightly slower convergence, and eventually can find solutions of very similar objective function values.

\section{Discussion}

\paragraph{Feedback Perturbation and Spike-Driven Learning.}
Our learning mechanism can be viewed as using the feedback connections to test the optimality of synaptic weights. 
As we have shown, an optimal network should receive little perturbation from feedback, and the derived learning rules correspond to local and greedy approaches to reduce the amount of drift in spike patterns.
Although our learning rules are based on spike rates, this idea certainly can be realized in a spike-driven manner to enable rapid correction of network dynamics.
In particular, spike timing dependent plasticity (STDP) is an ideal candidate to implement the feedforward and feedback learning rules.
The learning rules in (\ref{eq:ff_learning_rule}) share the same form with differential Hebbian and anti-Hebbian plasticity, whose link to STDP has been shown \cite{xie2000spike}.
On the other hand, the connection between our lateral learning rule and spike ordering based learning is less clear.
It can be seen that the rule is driven by shifts in postsynaptic spike rates, but a feasible mechanism to capture the exact weight dependency remains an open problem.

In autoencoder learning, \cite{hinton1988learning,burbank2015mirrored} similarly explored using feedback synapses for gradient computations.
However, the lack of lateral connectivities in an autoencoder makes it difficult to handle potential reverberation, and time delays are needed to separate the activities of the input and sparse-code (or hidden) layers. In contrast, our learning mechanism is based on the steady states of two network configurations. This strategy is actually a form of contrastive Hebbian learning \cite{movellan1991contrastive} in that the feedback synapses serve to bring the network from its ``free state'' to a ``clamped state''.

\paragraph{Practical Value.}

The proposed algorithm shows that the dictionary learning problem can be solved with fine-grained parallelism.
The synaptically local property means the computations can be fully distributed to individual neurons, eliminating the bottlenecking central unit.
The parallelism is best exploited by mapping the spiking network to a VLSI architecture, e.g., \cite{merolla2014million}, where each neuron can be implemented as a processing element.
Existing dictionary learning algorithms, e.g., \cite{aharon2006rm,mairal2009online}, can be accelerated by exploiting data parallelism, while it is less clear how to parallelize them within a single training sample to further reduce computation latency.

Our learning rules can be applied to related sparse coding models, such as reweighted $\ell_1$ minimization \cite{garrigues2010group} and Elastic Net \cite{ZouHastie05} (see \cite{charles2012common,tang2016sparse} for the respective dynamical system formulations).
It can also be extended to be a parallel solver for convolutional sparse coding \cite{zeiler2010deconvolutional,bristow2013fast}.
Although the weight sharing property in a convolutional model is fundamentally ``non-local'', this limitation may be overcame by clever memory lookup methods, as is commonly done in the computation of convolutional neural networks.

\section*{Acknowledgments}
The author thanks Peter Tang, Javier Turek, Narayan Srinivasa and Stephen Tarsa for insightful discussion and feedback on the manuscript, and Hong Wang for encouragement and support.

\small
\bibliographystyle{habbrv}
\bibliography{snn}

\end{document}